\documentclass[10pt, conference, compsocconf]{IEEEtran}

\usepackage{cite}
\usepackage{graphicx}
\usepackage{booktabs}
\usepackage{color}
\usepackage{amsmath}
\usepackage{amssymb}
\usepackage{algorithm}
\usepackage{algorithmic}

\usepackage{setspace}

\begin{document}
%
\title{Lecture video indexing using boosted margin maximizing neural networks}


\author{\IEEEauthorblockN{Di Ma\IEEEauthorrefmark{1},
Xi Zhang\IEEEauthorrefmark{2}, Xu Ouyang\IEEEauthorrefmark{3} and
Gady Agam\IEEEauthorrefmark{4}}
\IEEEauthorblockA{Department of Computer Science,
Illinois Institute of Technology\\
Chicago, Illinois, USA\\
Email: \IEEEauthorrefmark{1}dma2@hawk.iit.edu,
\IEEEauthorrefmark{2}xzhagn22@hawk.iit.edu,
\IEEEauthorrefmark{3}xouyang3@hawk.iit.edu,
\IEEEauthorrefmark{4}agam@iit.edu}}


\maketitle

\begin{abstract}
This paper presents a novel approach for lecture video indexing using a boosted deep convolutional neural network system. The indexing is performed by matching high quality slide images, for which text is either known or extracted, to lower resolution video frames with possible noise, perspective distortion, and occlusions. We propose a deep neural network integrated with a boosting framework composed of two sub-networks targeting feature extraction and similarity determination to perform the matching. The trained network is given as input a pair of slide image and a candidate video frame image and produces the similarity between them. A boosting framework is integrated into our proposed network during the training process. Experimental results show that the proposed approach is much more capable of handling occlusion, spatial transformations, and other types of noises when compared with known approaches.
\end{abstract}

\begin{IEEEkeywords}
Convolutional neural networks; Boosting algorithm; Lecture video indexing;

\end{IEEEkeywords}


\section{Introduction}


With recent year advancements, more and more information is recorded in multimedia documents instead of traditional documents. Specifically, classroom lectures are often recorded, shared and distributed as digital videos. In general, lectures span several hours and can produce large video files. This makes efficient access of content within lectures difficult. Automated video indexing allows quick retrieval of specific sections of interest without having to watch all the videos involved. Slides given by instructors can provide good indexes for lecture videos, if matched correctly to the videos. Specifically, searching text in slides can be used to query topics of interest from a large set of lecture videos, if the slides are linked to video segments. This association can be performed by matching high quality slide images to video frames. We assume that the high quality slide images are either OCRed or that their text is available. 

We use text image matching to link slide images to video frames. While the slide images and video frames have the same content, they may exhibit different perspective views, resolution, illumination, noise, and occlusions. Text image matching techniques are widely used in several areas besides matching slides to video frames for lecture video indexing \cite{di2014,di2015}. For example, when building digital libraries, numerous paper documents are scanned and archived as document images. It is important to ensure that duplicate pages are removed by text image matching techniques so that only one copy of a document exists in the library. This is done to facilitate accurate, non-redundant indexing and reducing storage cost. Postal automation \cite{liu1} is another good example for text image matching, where matching techniques can be used for identifying the same letter scanned by different letter-sorting machines. 

%

This paper has two novel contributions. First, we propose a boosted deep neural network system to perform slide matching for lecture video indexing. The proposed approach is much more resilient to occlusion, spatial transformations, and other types of noise compared with existing methods. Second, the proposed neural network system is trained as a classification margin maximizer by integrating boosting framework in the proposed training process. We show how such combination can efficiently train the proposed neural network system using just a small subset of samples from a huge number of samples.

\section{Related work}

During the past several decades, a variety of methods have been introduced to address the lecture video indexing problem. Yang et al. presented an automatic lecture video indexing algorithm by using video OCR technology in \cite{yang2011automatic}. Tuna et al. presented an approach for lecture video indexing in \cite{tuna2012development,adcock2010}. Global frame difference metrics were used for lecture video segmentation, which were then followed by OCR to extract textual metadata from slide frames and use it for indexing. In these two methods, the performance of segmentation and OCR strongly affects the quality of indexing.

Several approaches \cite{tuna2012development,adcock2010,CBMI2016} make use of global pixel-level-differencing metrics (e.g. HOG) for capturing slide transitions and matching slides to video frames to provide index points. A drawback of such approaches is that global differences can fail to generate accurate video segmentation results, when video frames have noise, or when slide frames change gradually. As a result, many redundant segments and indexes will be created. Jeong et al. proposed a lecture video indexing method using Scale Invariant Feature Transform (SIFT) features and the adaptive threshold \cite{jeong2012accurate}. In their work SIFT features are applied to measure slides with similar content, and an adaptive threshold selection algorithm is used to detect slide transitions. SIFT feature can handle various image transformations well, but may fail, when two images are from different sources.

\begin{figure*}[ht]
\centering
\includegraphics[scale=0.6]{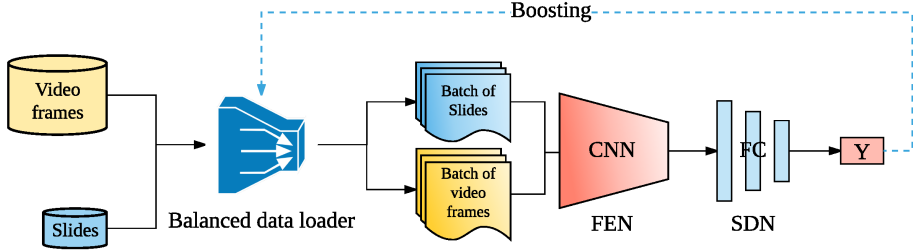} 
\caption{The architecture of the proposed system.}
\label{fig_flow}
\end{figure*}

Text image matching is a key part of lecture video indexing in the proposed approach. Lopresti \cite{lopr} applied Optical Character Recognition (OCR) to each text image, based on which approximate string matching techniques were used for near-duplicate document detection. However, OCR usually requires high quality images and is both language-dependent and time-consuming. Vitaladevuni et al. \cite{vita} employed interest points in detecting near-duplicate document images. Since there are often hundreds or thousands of interest points per image, this approach may exert a prohibitive computational overhead. Moreover, in document images with large amounts of noise, numerous incorrect corresponding points could be introduced and lead this type of algorithms to fail. About CNN based method, in \cite{adam2015}, the authors present a state-of-the-art method for document image classification and retrieval, using features learned by deep convolutional neural networks (CNNs). However, this method focuses on classifying documents into different categories. Retrieval is performed by retrieving files from specified classes. 

Several works related to image patch matching has been developed recently. Han et al. proposed a convolutional neural network, MatchNet \cite{matchnet}, designed for unifying feature and metric learning for patch-based matching. This system achieves state-of-the-art performance on a standard dataset for patch matching. However, the original MatchNet uses $64 \times 64$ small image patches as input. In our problem, downsampling slide and lecture video images to $64 \times 64$ loses too much text details. Specifically, the data sampling used in MatchNet causes slide images in one lecture video to be very similar with each other. Another popular method for image patch matching is the triplet network presented by Hoffer et al \cite{Hoffer2015}. In this method, three images are used as input including two images from a same class and a image from a different class. A metric that maximizes the distance between images from different classes is learned. The triplet net has comparable performance to the proposed approach which matching document images. However, training a triplet net is very tricky where the selection of three images as input plays an important role directly affecting both the convergence speed and the final quality of the trained network. Note that these methods are not designed for matching document images with identical text content.

Boosting is a general method for improving the accuracy of any given learning algorithm. Boosting has achieved a lot of success on different areas and applications. Its usage in training neural networks including deep neural networks can be traced back to many years ago. Bengio et al. \cite{schwenk2000boosting} investigated the performance of AdaBoost in neural networks and found that bootstrap resampling contained in the boosting framework was essential to obtain reduction in generalization error. In \cite{NIPS2016_6258} the authors integrate boosting with CNN by adding a boosting layer to incrementally select discriminative neurons from lower layers. They show that the performance of CNN with such a boosting layer is better than traditional CNN. Instead of training multiple weak learners, a successful usage of boosting in a single learner has been introduced in \cite{shalev2014selfieboost}. In \cite{karianakis2015boosting} boosting also helps learning deep representation of underlying data, where by using the boosting technique, a new manifold with better representation of features is discovered by combining features from different hierarchies of CNN. In the proposed approach, the boosting training framework as well as weighted sampling techniques are integrated into the training process. Instead of training multiple weak learners, a single network system is trained.
 
\section{The proposed approach}
In this paper, we employ a system based on a deep neural network with boosting framework for image retrieval by measuring image similarity. The proposed network system is inspired from the MatchNet \cite{matchnet}. Three modifications are made for solving our problem, including more layers for larger size of input image data, an additional feature concatenation layer and a different loss function. Furthermore, a boosting framework with a balance data loader is integrated into the proposed system. Figure \ref{fig_flow} shows the architecture of the proposed system.

\subsection{A two network system solution}
\label{sec: two network system}
To enable automatic feature extraction and similarity computation, the proposed system is composed of two neural networks that are connected to each other: a feature extraction network (FEN) and a similarity discrimination network (SDN). The FEN extracts features from input images and the SDN automatically computes a similarity score between a pair of image features. The two networks are trained as a whole system in an end-to-end manner. To train the system, we use training data composed of query images $Q = \{q_i\}$ and target images $T = \{t_i\}$, whereas there is at least one matching image for each $q_i\in Q$ in $T$. We approach the training of the system as a supervised training problem where in each training sample $(x_i, y_i)$, $x_i$ contains a pair of images $(q_i, t_i)$ randomly chosen from $Q$ and $T$, and $y_i$ is a $\{0, 1\}$ binary label representing whether the pair of images $(q_i, t_i)$ is a match. During training, we feed images $q_i$ and $t_i$ to the FEN to extract their features $f(q_i)$ and $f(t_i)$ respectively. The features $f(q_i)$ and $f(t_i)$ are then concatenated to form a longer feature vector $f(q_i)\oplus f(t_i)$. 

The order of $f(q_i)$ and $f(t_i)$ in the concatenation determines how the SDN is trained. Given any pair of image features, the SDN should be order-invariant. Thus, we create a symmetric feature vector  $f(t_i)\oplus f(q_i)$ and feed it to the SDN as well. An average error of these the two feature vectors is computed. We observe that retrieval results were improved when using this symmetric concatenation. We believe that this is because when the SDN is trained in an order-invariant manner, features extracted using the FEN are more robust and representative when used to distinguish a pair of images.



\subsection{Structure of the proposed network system}
The proposed system contains two networks, a feature extraction network (FEN) and a similarity discrimination network (SDN). The FEN is an 8 layers network consisting of 7 convolution layers and 1 fully connected layer. Batch normalization \cite{ioffe2015batch} followed by a leaky rectified linear unit (LReLU) is used for all convolution layers in FEN. Such configuration has been demonstrated to be both efficient and effective in many approaches \cite{kare,kriz,hint}. Kernels of $3\times 3$ size are used universally. There are 16 kernels at the first convolution layer. We double the number of kernels at each new layer, while at the same time, the size of images is halved using max-pooling. A full connection layer with 1024 nodes is stacked on top of the last convolution layer. A tanh activation is used for this layer. 

The SDN is a 3-layer fully connected neural network. The numbers of neurons in each layer are: 1024, 512 and 256 respectively. The tanh activation function is used for all layers except the last one where a linear output is used. This configuration will be discussed further in section \ref{sec: max margin}.

\subsection{Learning to approximately maximize the decision margin}
\label{sec: max margin}
Successfully training a well generalized neural network usually requires a large amount of training data. When it comes to the proposed two network system, given a training sample $(q_i, t_i)$ randomly picked from query image set $Q$ and target image set $T$, to cover all possible combination of samples in this problem, a training set contains $|Q| \times|T|$ samples should be generated. With such a large number of samples, the training process becomes very inefficient.

A feasible solution of this problem is to train only a randomly chosen subset from the overall training samples in each epoch. However, this solution will easily cause two problems in practice. First, by nature, our dataset is highly imbalanced, where the number of negative samples is substantially larger than the number of positive samples. This is because for each slide image, there are only a few matching frame images to form positive training samples. Training directly using imbalanced data like ours will cause a biased prediction towards negative cases\cite{Xi_CIKM_16}. Second, the decision boundary in a binary classification problem is often defined by samples that are close to the boundary. A well-known example is support vector machine' (SVM'), where the decision boundary is only defined by support vectors that lay on the margin. Thus, a blindly chosen training subset can result in a moving decision boundary that changes from iteration to iteration. This may cause slower convergence to a stable decision boundary.

To solve the problems mentioned above, in this paper, we propose to use a training framework similar to boosting algorithm, to maximize margins we use in addition a data loading balancer to solve the problems of imbalanced learning with a huge training dataset as described above. Details of these two approaches are described in following sections.

\subsubsection{Maximizing the margin by boosting}
Boosting is a well known technique for primarily reducing bias and also variance in supervised learning. Boosting consists of iteratively learning weak classifiers with respect to a distribution and adding them to a final strong classifier. When classifiers are added, they are typically weighted in some way that is usually related to the weak learners' accuracy. After a weak learner is added, the data are reweighted : examples that are misclassified gain weight and examples that are classified correctly loss weight. Thus future weak learners focus more on the examples that previous weak learners misclassified. 

Schapire et al.\cite{schapire1990strength} explain that the success of boosting is due to its property of increasing the margin. If the margin increases, the training instances are better separated and an error is less likely. This makes boosting's aim similar to that of support vector machines.

\begin{algorithm}[htb]  
\small
\caption{Boosting framework for training FEN-SDN system.}   
\label{Alg: BoostingFramwork}    	
\begin{algorithmic}[1]  	\REQUIRE ~~\\ 		
$\bullet$ A balance data loader L. \\ 	
$\bullet$ A hash table H\\ 	
$\bullet$ Training rounds R \\ 	
	\WHILE{R is not 0} 	
		\STATE Fetch $D = \{x_i=(q_i, t_i), y_i\}_{i = 1}^M$ from L. 		
		\STATE Get weights $\{W(x_i)\}_{i=1}^M$ from H.
		\STATE $\hat{W(x_i)} = \frac{W(x_i)}{\sum W(x_i)}$ for $1\leq i \leq M$
		\STATE Train FEN-SDN using $D$.
		\STATE Compute $\hat{y_i}$ as an output of FEN-SDN for each $x_i\in D$.
		\STATE Calculate error rate $\epsilon = \sum{\hat{W(x_i)}\cdot}$I$(y_i\neq \hat{y_i})$.	
		\STATE $\beta = \frac{1-\epsilon}{\epsilon}$.	
		\STATE For samples where $\hat(y_i) \neq y_i$, $W(x_i) = \beta\cdot W(x_i)$.
		\STATE Accordingly update weights for samples in H.
		\STATE $R = R - 1$	
	\ENDWHILE 
\end{algorithmic} 
\end{algorithm} 

The boosting framework adopted in this work is different from the original boosting framework, in that the weighted votes from weak learners are linearly combined as the final result. Thus we produce a single learner. By adopting the boosting framework, the algorithm only update the weights of samples and the final classification results is performed by a single network. 

For each training round in our boosting framework, our data loading balancer provides a dataset $D = \{x_i=(q_i, t_i), y_i\}_{i = 1}^m$ for the current training round. The proposed network system is trained by $K$ epochs with mini-batch of data extracted from $D$ in each epoch. Given the fact that there are $|Q| \times |T|$ possible samples in total, $D$ is just a small portion that is visible to the proposed network system in each round. We don't keep track the weights for all possible samples for the sake of saving memory. Instead, we use a hash table H to save the weights of all samples that are trained so far. Thus, it is possible that all weights that are currently available in the hash table do not add up to one, and so there is a need to normalize the weights for all samples extracted for training in the current round. For training samples that are used the first time, we initialize their weights as $W(x_i) = \frac{1}{|Q|\times |T|}$. Once training is done for samples in the current round, we update their weights in the hash table $H$ accordingly. 

The proposed training framework using boosting is summarized by Algorithm \ref{Alg: BoostingFramwork}.  

\subsubsection{A balanced data loader}
\label{data loader}
Normally, for a relatively small dataset, it is feasible to load all samples and keep them in memory. However, when a dataset contains a huge number of samples such as the one used in this work ($|Q|\times |T|$). Reading all samples may not be possible. In addition, there are more negative samples where a query image does not match with a target image than positive samples. To have a balanced data for training an unbiased classifier, we use a mechanism in this work, a data loader,  which works as a separate process to pre-fetch samples used to train the proposed network system in each round.

The data loader in this work is responsible for generating a training set based on the current weights of samples in the hash table with a balanced class membership distribution. The inputs to the data loader are the query image set $Q$, target image set $T$, and the hash table $H$. Suppose that there are on average $\alpha$ (where $\alpha \geq 1$) matching targets for each query image. To enable an equal number of drawings of positive and negative samples from entire dataset, the probabilities of picking positive samples and negative samples are set to  $P_{+} = 1-\frac{\alpha}{|T|}$ and $P_{-}=\frac{\alpha}{|T|}$ respectively. We consider samples with weights ranked in the top $\mu$ percentage as hard samples which are samples potentially laying close to the classification margin. After the data loading process, a dataset $D=\{x_i = (q_i, t_i), y_i\}_{i=1}^m$ is provided by data loader in each training round. 

We summarize details of loading data in Algorithm \ref{Alg: data loader}.

\begin{algorithm}[htb]  
\small
\caption{Fetch data using data loader.}   
\label{Alg: data loader}    	
\begin{algorithmic}[1]  	\REQUIRE ~~\\ 		
$\bullet$ A hash table H.\\ 	
$\bullet$ Query set $Q$ and target set $T$.\\ 	
$\bullet$ Probability of drawing positive and negative samples: $P_{+}$ and $P_{-}$.\\
$\bullet$ Total number samples in the output $m$.\\
$\bullet$ Hard case rate $\mu$.\\
\WHILE{$|D|\leq m$} 	
	\STATE Randomly pick $k$ query images and $k$ target images. Then pair them up. Get $K=\{x_i = (q_i, t_i)), y_i\}_{i=1}^k$
	\STATE Check weights from hash table T for samples in $K$. If new to T, assign $W(x_i)=\frac{1}{|Q|\times|T|}$.
	\STATE Draw an equal number of positive and negative cases with top $\mu$ percentage of weights from $K$. Denoted as $K_{\mu}$.
	\STATE $D = D \cup K_{\mu}$ 
\ENDWHILE 
\STATE Return D.
\end{algorithmic} 
\end{algorithm} 

\subsubsection{Hinge loss}
By maximizing the decision margin, not only do we require that samples are classified correctly, but also we desire that correctly classified samples lay far away from the decision boundary. Such design strategy is essentially as same as that of SVM where hinge loss is used to maximize the margin. Thus, we use hinge loss as the objective function when training our FEN-SDN network system. Given $y_i$ as a ground truth label of sample $x_i$ and $\hat{y_i}$ the classifier output score, the hinge loss of the prediction $\hat(y_i)$ is defined as:
\begin{equation}
E(x_i) = max(0, 1-y_i \cdot \hat{y_i})
\end{equation}

Since assuming that class labels are $+1$ or$-1$, hinge loss is not differentiable, it will fail in gradient descent optimization. Hence, we employ the squared hinge loss from L2-SVM which penalizes violated margins more strongly (quadratically instead of linearly). The squared hinge loss has the form:
\begin{equation}
E(x_i) = \frac{1}{2\delta} max(0, 1-y_i \cdot \hat{y_i})^2
\end{equation}

where $\delta$ is a smoothing parameter (e.g. $\delta = 2$).

\subsection{Training of the network}
The proposed neural network system is trained using mini-batch. Since the size of an input image is $256\times 256$, to avoid memory overflow, we use a relatively small number of image pairs in each mini batch and so the batch size is set to 32. The network is trained using adaptive moment estimation (Adam) with learning rate set to $0.0002$ and momentum set to $0.5$.  The training is also regularised by weight decay (the $L_2$ penalty multiplier is set to $5\cdot 10^{-4}$). We initialize the network weights using the approach proposed by He \cite{he2015delving}. 


\section{Experimental evaluation}
In this section, we provide experimental evaluation of the proposed approach. We first introduce the generation of the dataset and then describe three benchmark methods \cite{vita,matchnet,Hoffer2015}, to which the results of our proposed approach can be compared.

\begin{figure}[ht]
\centering
\includegraphics[scale=0.08]{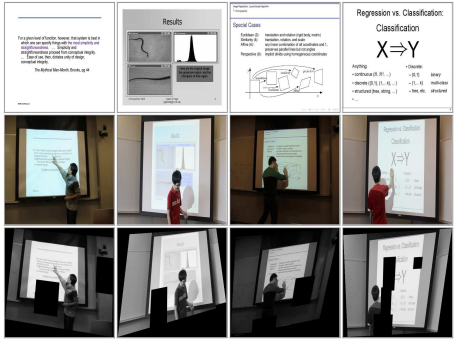} 
\caption{Examples of the dataset include per row: target slide images, real query video frames, and synthetic query frames.}
\label{fig_trainingdata}
\end{figure}

\subsection{Dataset}
To the best of our knowledge, there is no publicly available dataset for the task of lecture video indexing. To generate a dataset we collected 150 videos associated with their course slides from the web. A video segmentation algorithm \cite{di2015} was applied to break the lecture videos into segments. We manually matched each slide with its corresponding video segment to create positive ground truth cases. All other possible combinations between slides and video frames were given negative labels.

As a result, for training, $3,000$ slide images were collected. Each slide is associated with 5 video frames. Theoretically, there are $15,000$ positive image pairs and $3000 \times (5 \times 2999) = 44,985,000$ negative image pairs. The amount of positive cases is much smaller and the dataset is very imbalanced. To augment the dataset, synthetic query video frame were generated by adding random black block noise and applying a random projective transformation to warp each video frame image.  We controlled the parameters of the projective transformation matrix in a reasonable ranges (e.g. scale: $[0.9,1.1]$, rotation: $[-30,30]$, translation: $[-50,50]$, etc.) so that a realistic transformation is produced. We also used 10 different levels of noise to generate the synthetic data. The noise level indicates the ratio of blocked pixels and the total number of pixels in the image. The number of the noise blocks  was also changed. Similar to the real data, each unique target slide image was paired with 5 positive synthetic query frame images. This process provided us extra $15,000$ positive synthetic image pairs and $3000 \times (5 \times 2999) = 44,985,000$ negative image pairs. The proposed data loader was responsible for creating a balanced dataset during training as described before. The testing dataset was created using $1,000$ slide images. Similar to the training data, each target slide image was associated with 5 video frames. The testing dataset includes $5,000$ positive image pairs and $5,000$ negative image pairs. All images were resized to $256 \times 256$ pixels before being fed into the proposed deep neural network system for learning.

\begin{table*}[]
\centering
\caption{Video frame retrieval results using 10 different quality levels. The results of our proposed method are compared to \cite{vita,matchnet,Hoffer2015} using top 1, 5, 10 hit rates.}
\begin{tabular}{ccccccccccccc}
\hline
      & \multicolumn{4}{c}{Top 1}                                & \multicolumn{4}{c}{Top 5}                                            & \multicolumn{4}{c}{Top 10}                                           \\ \hline
Quality & Proposed & \multicolumn{1}{c}{\cite{Hoffer2015}} & \multicolumn{1}{c}{\cite{matchnet}} & \multicolumn{1}{c}{\cite{vita}} &
\multicolumn{1}{c}{Proposed} & \multicolumn{1}{c}{\cite{Hoffer2015}} & \multicolumn{1}{c}{\cite{matchnet}} & \multicolumn{1}{c}{\cite{vita}} &
\multicolumn{1}{c}{Proposed} & \multicolumn{1}{c}{\cite{Hoffer2015}} & \multicolumn{1}{c}{\cite{matchnet}} & \multicolumn{1}{c}{\cite{vita}}\\ \hline

1 & \textbf{0.932} & 0.929 & 0.912 & 0.846 & \textbf{0.981} & 0.975 & 0.972 & 0.961 & \textbf{0.981} & 0.981 & 0.987 & 0.986\\
2 & 0.913 & \textbf{0.914} & 0.895 & 0.813 & \textbf{1.000} & 0.974 & 0.955 & 0.936 & \textbf{1.000} & 0.985 & 0.985 & 0.967\\
3 & \textbf{0.888} & 0.883 & 0.853 & 0.747 & 0.969 & \textbf{0.970} & 0.941 & 0.892 & \textbf{0.993} & 0.985 & 0.985 & 0.934\\
4 & \textbf{0.888} & 0.870 & 0.832 & 0.664 & \textbf{0.987} & 0.946 & 0.936 & 0.985 & \textbf{0.993} & 0.985 & 0.983 & 0.884\\
5 & 0.851 & \textbf{0.855} & 0.806 & 0.543 & \textbf{0.981} & 0.946 & 0.932 & 0.973 & \textbf{0.993} & 0.980 & 0.983 & 0.826\\
6 & \textbf{0.771} & 0.765 & 0.754 & 0.458 & 0.930 & \textbf{0.950} & 0.881 & 0.750 & 0.938 & \textbf{0.940} & 0.932 & 0.782\\
7 & \textbf{0.765} & 0.750 & 0.738 & 0.436 & \textbf{0.913} & 0.890 & 0.878 & 0.642 & \textbf{0.950} & 0.948 & 0.894 & 0.744\\
8 & \textbf{0.740} & \textbf{0.740} & 0.707 & 0.456 & \textbf{0.913} & 0.884 & 0.874 & 0.634 & \textbf{0.944} & 0.946 & 0.880 & 0.730\\
9 & 0.648 & \textbf{0.670} & 0.603 & 0.414 & \textbf{0.870} & 0.846 & 0.845 & 0.596 & 0.913 & \textbf{0.920} & 0.848 & 0.678\\
10 & \textbf{0.592} & 0.550 & 0.553 & 0.376 & \textbf{0.790} & 0.781 & 0.785 & 0.596 & \textbf{0.858} & 0.830 & 0.828 & 0.689\\
all & \textbf{0.905} & 0.885 & 0.863 & 0.722 & \textbf{0.985} & 0.948 & 0.945 & 0.833 & \textbf{0.992} & 0.985 & 0.988 & 0.902\\\hline

\end{tabular}

\label{table_compare}
\end{table*}

\subsection{Benchmarks}
We compared three methods to the proposed system. The first method is a traditional interest point based approach\cite{vita}. This benchmark algorithm computes SIFT interest point features\cite{Lowe2004} on a set of images to build a database. Given a query image, its SIFT features are extracted and their nearest neighbours in the feature database are retrieved. The second and third methods are convolutional network based approaches: MatchNet\cite{matchnet} and the triplet net\cite{Hoffer2015}. Since the original MatchNet can only take $64 \times 64$ images as input, we added two additional convolutional layers to adapt the input images. These convolutional neural networks extract features from image patches and compute similarities between them.

\begin{figure}[ht]
\centering
\includegraphics[scale=0.4]{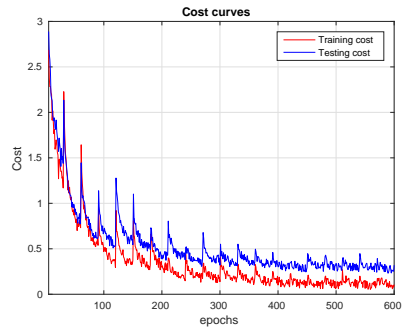} 
\caption{Training and testing costs of the proposed network system.}
\label{fig_costs}
\end{figure}

\subsection{Experimental results}
%
%
%
\begin{figure}[ht]
\centering
\includegraphics[scale=0.075]{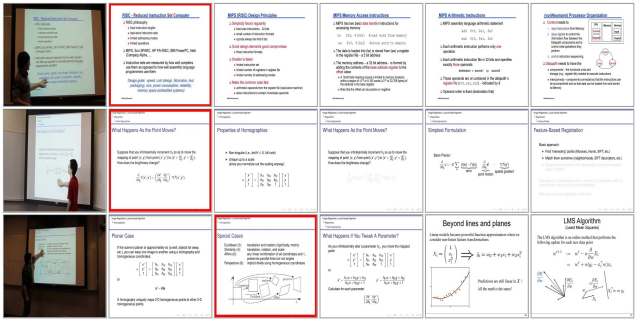} 
\caption{Retrieval results of query video frame images. The query frames are shown in the left column. Top 5 retrieval results are listed in a descending order according to matching scores produced by the proposed neural network system. The correct matches are highlighted in red.}
\label{fig_query1}
\end{figure}
%

\subsubsection{Performance of the proposed method for lecture video index frame retrieval}

We first evaluate the proposed deep neural network system for lecture video index frame retrieval at different noise levels. The proposed deep neural network system generates a matching score for each image pair. The matching scores are sorted and the best candidates of matched slide images are selected from the target slide image set. We compare the proposed approach with the three benchmark methods \cite{vita,matchnet,Hoffer2015}. Taking into account that slides on video frames have variable quality caused by occlusion, lighting conditions and image deformation due to different viewing angles, we evaluate the proposed neural network system using various video images with different qualities. We define image quality as a measure of size of slide area, blocked area, text area and the extent of perspective distortion. We employed the approaches proposed in \cite{yang2011automatic,imran2016} to detect text areas, slide areas and human foreground. The overall image quality is calculated by:
\begin{equation}
Q_i = T_i \times S_i \times (C - B_i)  
\end{equation}

In this equation, $Q_i$ is the quality value of image $i$. $C$ is the size of image $i$, which is a constant $256\times256$ in our experiments. $T_i$, $S_i$ and $B_i$ are the text area, slide area and blocked area respectively. Based on the image quality value, we divide our query video frames into 10 levels. Video indexing tests are conducted on query images of each level separately and the top 1, top 5 and top 10 hit rates are summarized in Table \ref{table_compare}.

As can be observed in Table \ref{table_compare}, the proposed approach has better performance compared with the benchmark methods, where the proposed approach achieves the highest hit rate on all experiments. It also shows that our proposed approach can handle noisy query frame image better than the benchmark methods. Note that besides the key component text area, other factors like image warping parameters, illumination, resolution, can also effect the performance and so it is possible that in some instances a set with a higher quality is more difficult to match and will produce a slightly lower hit rate. Figure \ref{fig_query1} shows the results of video frame retrieval. The first image of each row is the query image. The subsequent five images are retrieval results sorted based on the matching score generated by our proposed neural network system. Correct target images are found in all four cases. We can also observe that when noise level increases, the accuracy decreases.

\section{Conclusions}

This paper presents a novel approach for lecture video indexing using boosted deep convolutional neural networks (CNNs). The indexing is performed by matching high quality slide images to lower resolution video frames with possible noise, perspective distortion, and occlusions for which text is either known or extracted.The proposed deep neural network system consists of two sub-networks targeting feature extraction and similarity determination to perform image matching for lecture video indexing. The trained network system is given as input a pair of slide image and a candidate video frame image and produces the similarity between them. A boosting framework is integrated into our proposed network during the training process. Experimental results show that the proposed approach is much more capable of handling occlusion, spatial transformations and other types of noises when compared with known approaches.






\bibliographystyle{./IEEEtran}
\bibliography{./IEEEabrv,./reference.bib}
%

%
%

\end{document}